# Hilbert Space Embeddings of Predictive State Representations

**Byron Boots**
Computer Science and Engineering Dept.
University of Washington
Seattle, WA

**Arthur Gretton**
Gatsby Unit
University College London
London, UK

**Geoffrey J. Gordon**
Machine Learning Dept.
Carnegie Mellon University
Pittsburgh, PA

## Abstract

Predictive State Representations (PSRs) are an expressive class of models for controlled stochastic processes. PSRs represent state as a set of predictions of future observable events. Because PSRs are defined entirely in terms of observable data, statistically consistent estimates of PSR parameters can be learned efficiently by manipulating moments of observed training data. Most learning algorithms for PSRs have assumed that actions and observations are finite with low cardinality. In this paper, we generalize PSRs to infinite sets of observations and actions, using the recent concept of Hilbert space embeddings of distributions. The essence is to represent the state as one or more nonparametric *conditional embedding operators* in a Reproducing Kernel Hilbert Space (RKHS) and leverage recent work in kernel methods to estimate, predict, and update the representation. We show that these *Hilbert space embeddings of PSRs* are able to gracefully handle continuous actions and observations, and that our learned models outperform competing system identification algorithms on several prediction benchmarks.

## 1 INTRODUCTION

Many problems in machine learning and artificial intelligence involve discrete-time partially observable nonlinear dynamical systems. If the observations are discrete, then *Hidden Markov Models* (HMMs) [1] or, in the control setting, *Input-Output HMMs* (IO-HMMs) [2], can be used to represent belief as a discrete distribution over latent states. *Predictive State Representations (PSRs)* [3] are generalizations of IO-HMMs that have attracted interest because they can have greater representational capacity for a fixed model dimension. In contrast to latent-variable representations like HMMs, PSRs represent the state of a dynamical system by tracking occurrence probabilities of future observable events (called *tests*) conditioned on past observable events (called *histories*). One of the prime motivations for modeling dynamical systems with PSRs was that, because tests and histories are observable quantities, learning PSRs should be easier than learning IO-HMMs by heuristics like Expectation Maximization (EM), which suffer from bad local optima and slow convergence rates.

For example, Boots et al. [4] proposed a *spectral* algorithm for learning PSRs with discrete observations and actions. At its core, the algorithm performs a singular value decomposition of a matrix of joint probabilities of tests and partitions of histories (the moments mentioned above), and then uses linear algebra to recover parameters that allow predicting, simulating, and filtering in the modeled system. As hinted above, the algorithm is statistically consistent, and does not need to resort to local search—an important benefit compared to typical heuristics (like EM) for learning latent variable representations.

Despite their positive properties, many algorithms for PSRs are restricted to discrete observations and actions with only moderate cardinality. For continuous actions and observations, and for actions and observations with large cardinalities, learning algorithms for PSRs often run into trouble: we cannot hope to see each action or observation more than a small number of times, so we cannot gather enough data to estimate the PSR parameters accurately without additional assumptions. Previous approaches attempt to learn continuous PSRs by leveraging kernel density estimation [4] or modeling PSR distributions with exponential families [5, 6]; each of these methods must contend with drawbacks such as slow rates of statistical convergence and difficult numerical integration.

In this paper, we fully generalize PSRs to continuous observations and actions using a recent concept called Hilbert space embeddings of distributions [7, 8].

The essence of our method is to represent distributions of tests, histories, observations, and actions as points in (possibly) infinite-dimensional reproducing kernel Hilbert spaces. During filtering we update these embedded distributions using a kernel version of Bayes' rule [9]. The advantage of this approach is that embedded distributions can be estimated accurately without having to contend with problems such as density estimation and numerical integration. Depending on the kernel, the model can be parametric or nonparametric. We focus on the nonparametric case: we leverage the "kernel trick" to represent the state and required operators implicitly and maintain a state vector with length proportional to the size of the training dataset.

### 1.1 RELATED WORK

Our approach is similar to recent work that applies kernel methods to dynamical system modeling and reinforcement learning, which we summarize here. Song et al. [10] proposed a nonparametric approach to learning HMM representations in RKHSs. The resulting dynamical system model, called Hilbert Space Embeddings of Hidden Markov Models (HSE-HMMs), proved to be more accurate compared to competing models on several experimental benchmarks [10, 11]. Despite these successes, HSE-HMMs have two major limitations: first, the update rule for the HMM relies on *density estimation* instead of Bayesian inference in Hilbert space, which results in an awkward model with poor theoretical guarantees. Second, the model lacks the capacity to reason about actions, which limits the scope of the algorithm. Our model can be viewed as an extension of HSE-HMMs that adds inputs and updates state using a kernelized version of Bayes' rule.

Grünewälder et al. [12] proposed a nonparametric approach to learning transition dynamics in Markov decision processes (MDPs) by representing the stochastic transitions as conditional distributions in RKHS. This work was extended to POMDPs by Nishiyama et al. [13]. Like the approach we propose here, the resulting Hilbert space embedding of POMDPs represents distributions over the states, observations, and actions as embeddings in RKHS and uses kernel Bayes' rule to update these distribution embeddings. Critically, the algorithm requires training data that includes labels for the true latent states. This is a serious limitation: it precludes learning dynamical systems directly from sensory data. By contrast, our algorithm only requires access to an unlabeled sequence of actions and observations, and learns the more expressive PSR model, which includes POMDPs as a special case.

## 2 PSRS

A PSR represents the state of a controlled stochastic process as a set of predictions of observable experiments or *tests* that can be performed in the system. Specifically, a test of length $N$ is an ordered sequence of future action-observations pairs $\tau = a_1, o_1, \ldots a_N, o_N$ that can be selected and observed at any time $t$. Likewise, a *history* is an ordered sequence of actions and observations $h = a_1, o_1, \ldots, a_M, o_M$ that have been selected and observed prior to $t$.

A test $\tau_i$ is *executed* at time $t$ if we intervene [14] to select the sequence of actions specified by the test $\tau_i^{\mathcal{A}} = a_1, \ldots, a_N$. It is said to *succeed* at time $t$ if it is executed and the sequence of observations in the test $\tau_i^{\mathcal{O}} = o_1, \ldots, o_N$ matches the observations generated by the system. The *prediction* for test $i$ at time $t$ is the probability of the test succeeding given a history $h_t$ and given that we execute it:[1]

$$\mathbb{P}\left[\tau_{i,t}^{\mathcal{O}} \mid \tau_{i,t}^{\mathcal{A}}, h_t\right] = \frac{\mathbb{P}\left[\tau_i^{\mathcal{O}}, \tau_i^{\mathcal{A}} \mid h_t\right]}{\mathbb{P}\left[\tau_i^{\mathcal{A}} \mid h_t\right]} \quad (1)$$

The key idea behind a PSR is that if we know the expected outcomes of executing *all* possible tests, then we know everything there is to know about the state of a dynamical system [16]. In practice we will work with the predictions of some *set* of tests; therefore, let $\mathcal{T} = \{\tau_i\}$ be a set of $d$ tests. We write

$$s(h_t) = \left(\mathbb{P}\left[\tau_{i,t}^{\mathcal{O}} \mid \tau_{i,t}^{\mathcal{A}}, h_t\right]\right)_{i=1}^{d} \quad (2)$$

for the *prediction vector* of success probabilities for the tests $\tau_i \in \mathcal{T}$ given a history $h_t$.

Knowing the success probabilities of some tests may allow us to compute the success probabilities of other tests. That is, given a test $\tau_l$ and a prediction vector $s(h_t)$, there may exist a *prediction function* $f_{\tau_l}$ such that $\mathbb{P}\left[\tau_l^{\mathcal{O}} \mid \tau_l^{\mathcal{A}}, h_t\right] = f_{\tau_l}(s(h_t))$. In this case, we say $s(h_t)$ is a *sufficient statistic* for $\mathbb{P}\left[\tau_l^{\mathcal{O}} \mid \tau_l^{\mathcal{A}}, h_t\right]$. A *core* set of tests is a set whose prediction vector $s(h_t)$ is a sufficient statistic for the predictions of *all* tests $\tau_l$ at time $t$. Therefore, $s(h_t)$ is a *state* for our PSR: i.e., at each time step $t$ we can remember $s(h_t)$ instead $h_t$.

Formally, a PSR is a tuple $\langle \mathcal{O}, \mathcal{A}, \mathcal{T}, \mathcal{F}, s_o \rangle$. $\mathcal{O}$ is the set of possible observations and $\mathcal{A}$ is the set of possible actions. $\mathcal{T}$ is a core set of tests. $\mathcal{F}$ is the set of prediction functions $f_{\tau_l}$ for *all* tests $\tau_l$ (which must exist since $\mathcal{T}$ is a core set), and $s_0 = s(h_0)$ is the initial prediction vector after seeing the empty history $h_0$.

In this paper we restrict ourselves to *linear* PSRs, in which all prediction functions are linear: $f_{\tau_l}(s(h_t)) =$

---

[1] For simplicity, we assume that all probabilities involving actions refer to our PSR as controlled by an arbitrary *blind* or *open-loop* policy [15] (also called *exogenous inputs*). In this case, conditioning on $\text{do}(a_1, \ldots, a_M)$ is equivalent to conditioning on observing $a_1, \ldots, a_M$, which allows us to avoid some complex notation and derivations.

$f_{\tau_l}^\mathsf{T} s(h_t)$ for some vector $f_{\tau_l} \in \mathbb{R}^{|\mathcal{T}|}$. Note that the restriction to linear prediction functions is only a restriction to linear relationships between conditional probabilities of tests; linear PSRs can still represent systems with *nonlinear* dynamics.

## 2.1 FILTERING WITH BAYES' RULE

After taking action $a$ and seeing observation $o$, we can update the state $s(h_t)$ to the state $s(h_{t+1})$ by Bayes' rule. The key idea is that the set of functions $\mathcal{F}$ allows us to predict *any* test from our core set of tests.

The state update proceeds as follows: first, we predict the success of any core test $\tau_i$ prepended by an action $a$ and an observation $o$, which we call $ao\tau_i$, as a linear function of our core test predictions $s(h_t)$:

$$\mathbb{P}\left[\tau_{i,t+1}^{\mathcal{O}}, o_t{=}o \mid \tau_{i,t+1}^{\mathcal{A}}, a_t{=}a, h_t\right] = f_{ao\tau_i}^\mathsf{T} s(h_t) \quad (3)$$

Second, we predict the likelihood of any observation $o$ given that we select action $a$ (i.e., the test $ao$):

$$\mathbb{P}\left[o_t = o \mid a_t = a, h_t\right] = f_{ao}^\mathsf{T} s(h_t) \quad (4)$$

After executing action $a$ and seeing observation $o$, Equations 3 and 4 allow us to find the prediction for a core test $\tau_i$ from $s(h_t)$ using Bayes' Rule:

$$\begin{aligned}
s_i(h_{t+1}) &= \mathbb{P}\left[\tau_{i,t+1}^{\mathcal{O}} \mid \tau_{i,t+1}^{\mathcal{A}}, a_t = a, o_t = o, h_t\right] \\
&= \frac{\mathbb{P}\left[\tau_{i,t+1}^{\mathcal{O}}, o_t = o \mid \tau_{i,t+1}^{\mathcal{A}}, a_t = a, h_t\right]}{\mathbb{P}\left[o_t = o \mid a_t = a, h_t\right]} \\
&= \frac{f_{ao\tau_i}^\mathsf{T} s(h_t)}{f_{ao}^\mathsf{T} s(h_t)}
\end{aligned} \quad (5)$$

This recursive application of Bayes' rule to a belief state is called a *Bayes filter*.

## 3 HILBERT SPACE EMBEDDINGS

The key idea in this paper is to represent (possibly continuous) distributions of tests, histories, observations, and actions *nonparametrically* as points in (possibly infinite dimensional) Hilbert spaces. During filtering these points are updated entirely in Hilbert space, mirroring the finite-dimensional updates, using a kernel version of Bayes' rule.

## 3.1 MEAN MAPS

Let $\mathcal{F}$ be a reproducing kernel Hilbert space (RKHS) associated with kernel $K_X(x, x') \stackrel{\text{def}}{=} \langle \phi^X(x), \phi^X(x') \rangle_{\mathcal{F}}$ for $x \in \mathcal{X}$. Let $\mathcal{P}$ be the set of probability distributions on $\mathcal{X}$, and $X$ be a random variable with distribution $\mathbb{P} \in \mathcal{P}$. Following Smola et al. [7], we define the mean map (or the embedding) of $\mathbb{P} \in \mathcal{P}$ into RKHS $\mathcal{F}$ to be $\mu_X \stackrel{\text{def}}{=} \mathbb{E}\left[\phi^X(X)\right]$.

A *characteristic* RKHS is one for which the mean map is injective: that is, each distribution $\mathbb{P}$ has a unique embedding [8]. This property holds for many commonly used kernels, e.g., the Gaussian and Laplace kernels when $\mathcal{X} = \mathbb{R}^d$.

Given *i.i.d.* observations $x_t$, $t = 1 \ldots T$, an estimate of the mean map is straightforward:

$$\hat{\mu}_X \stackrel{\text{def}}{=} \frac{1}{T} \sum_{t=1}^T \phi^X(x_t) = \frac{1}{T} \Upsilon^X \mathbf{1}_T \quad (6)$$

where $\Upsilon^X \stackrel{\text{def}}{=} \left(\phi^X(x_1), \ldots, \phi^X(x_T)\right)$ is the linear operator which maps the $t$th unit vector of $\mathbb{R}^T$ to $\phi^X(x_t)$.

Below, we'll sometimes need to embed a joint distribution $\mathbb{P}[X, Y]$. It is natural to embed $\mathbb{P}[X, Y]$ into a tensor product RKHS: let $K_Y(y, y') = \langle \phi^Y(y), \phi^Y(y') \rangle_{\mathcal{G}}$ be a kernel on $\mathcal{Y}$ with associated RKHS $\mathcal{G}$. Then we write $\mu_{XY}$ for the mean map of $\mathbb{P}[X, Y]$ under the kernel $K_{XY}((x, y), (x', y')) \stackrel{\text{def}}{=} K_X(x, x') K_Y(y, y')$ for the tensor product RKHS $\mathcal{F} \otimes \mathcal{G}$.

## 3.2 COVARIANCE OPERATORS

The covariance operator is a generalization of the covariance matrix. Given a joint distribution $\mathbb{P}[X, Y]$ over two variables $X$ on $\mathcal{X}$ and $Y$ on $\mathcal{Y}$, the *uncentered* covariance operator $\mathcal{C}_{XY}$ is the linear operator which satisfies [17]

$$\langle f, \mathcal{C}_{XY} g \rangle_{\mathcal{F}} = \mathbb{E}_{XY}\left[f(X) g(Y)\right] \quad \forall f \in \mathcal{F}, g \in \mathcal{G} \quad (7)$$

Both $\mu_{XY}$ and $\mathcal{C}_{XY}$ represent the distribution $\mathbb{P}[X, Y]$. One is defined as an element of $\mathcal{F} \otimes \mathcal{G}$, and the other as a linear operator from $\mathcal{G}$ to $\mathcal{F}$, but they are isomorphic under the standard identification of these spaces [9], so we abuse notation and write $\mu_{XY} = \mathcal{C}_{XY}$.

Given $T$ *i.i.d.* pairs of observations $(x_t, y_t)$, define $\Upsilon^X = \left(\phi^X(x_1), \ldots, \phi^X(x_T)\right)$ and $\Upsilon^Y = \left(\phi^Y(y_1), \ldots, \phi^Y(y_T)\right)$. Write $\Upsilon^*$ for the adjoint of $\Upsilon$. Analogous to (6), we can estimate

$$\widehat{\mathcal{C}}_{XY} = \frac{1}{T} \Upsilon^X \Upsilon^{Y*} \quad (8)$$

## 3.3 CONDITIONAL OPERATORS

Based on covariance operators, Song et al. [18] define a linear operator $\mathcal{W}_{Y|X} : \mathcal{F} \mapsto \mathcal{G}$ that allows us to compute conditional expectations $\mathbb{E}\left[\phi^Y(Y) \mid x\right]$ in RKHSs. Given some smoothness assumptions [18], this conditional embedding operator is

$$\mathcal{W}_{Y|X} \stackrel{\text{def}}{=} \mathcal{C}_{YX} \mathcal{C}_{XX}^{-1} \quad (9)$$

and for all $g \in \mathcal{G}$ we have

$$\mathbb{E}[g(Y) \mid x] = \langle g, \mathcal{W}_{Y|X} \phi^X(x) \rangle_{\mathcal{G}}$$

Given $T$ *i.i.d.* pairs $(x_t, y_t)$ from $\mathbb{P}[X, Y]$, we can estimate $\mathcal{W}_{Y|X}$ by kernel ridge regression [18, 19]:

$$\widehat{\mathcal{W}}_{Y|X} = (1/T) \Upsilon^Y \left((1/T) \Upsilon^X\right)_\lambda^\dagger$$

where the regularized pseudoinverse $\Upsilon_\lambda^\dagger$ is given by $\Upsilon_\lambda^\dagger = \Upsilon^*(\Upsilon\Upsilon^* + \lambda I)^{-1}$. (The regularization parameter $\lambda$ helps to avoid overfitting and to ensure invertibility, and thus that the resulting operator is well defined.) Equivalently,

$$\widehat{\mathcal{W}}_{Y|X} = \Upsilon^Y(G_{X,X} + \lambda TI)^{-1}\Upsilon^{X*}$$

where the Gram matrix $G_{X,X} \stackrel{\text{def}}{=} \Upsilon^{X*}\Upsilon^X$ has $(i,j)$th entry $K_X(x_i, x_j)$.

### 3.4 KERNEL BAYES' RULE

We are now in a position to define the kernel mean map implementation of Bayes' rule (called the Kernel Bayes' Rule, or KBR). In particular, we want the kernel analog of $\mathbb{P}[X \mid y, z] = \mathbb{P}[X, y \mid z] / \mathbb{P}[y \mid z]$. In deriving the kernel realization of this rule we need the kernel mean representation of a conditional *joint* probability $\mathbb{P}[X, Y \mid z]$. Given Hilbert spaces $\mathcal{F}$, $\mathcal{G}$, and $\mathcal{H}$ corresponding to the random variables $X$, $Y$, and $Z$ respectively, $\mathbb{P}[X, Y \mid z]$ can be represented as a mean map $\mu_{XY|z} \stackrel{\text{def}}{=} \mathbb{E}\left[\phi^X(X) \otimes \phi^Y(Y) \mid z\right]$ or the corresponding operator $\mathcal{C}_{XY|z}$. Under some assumptions [9], and with a similar abuse of notation as before, this operator satisfies:

$$\mathcal{C}_{XY|z} = \mu_{XY|z} \stackrel{\text{def}}{=} \mathcal{C}_{(XY)Z}\mathcal{C}_{ZZ}^{-1}\phi(z) \qquad (10)$$

Here the operator $\mathcal{C}_{(XY)Z}$ represents the covariance of the random variable $(X, Y)$ with the random variable $Z$. (We can view (10) as applying a conditional embedding operator $\mathcal{W}_{XY|Z}$ to an observation $z$.) We now define KBR in terms of conditional covariance operators [9]:

$$\mu_{X|y,z} = \mathcal{C}_{XY|z}\mathcal{C}_{YY|z}^{-1}\phi(y) \qquad (11)$$

To map the KBR to the ordinary Bayes' rule above, $\mu_{X|y,z}$ is the embedding of $\mathbb{P}[X \mid y, z]$; $\mathcal{C}_{XY|z}$ is the embedding of $\mathbb{P}[X, Y \mid z]$; and the action of $\mathcal{C}_{YY|z}^{-1}\phi(y)$ corresponds to substituting $Y = y$ into $\mathbb{P}[X, Y \mid z]$ and dividing by $\mathbb{P}[y \mid z]$.

To use KBR in practice, we need to estimate the operators on the RKHS of (11) from data. Given $T$ i.i.d. triples $(x_t, y_t, z_t)$ from $\mathbb{P}[X, Y, Z]$, write $\Upsilon^X = \left(\phi^X(x_1), \ldots, \phi^X(x_T)\right)$, $\Upsilon^Y = \left(\phi^Y(y_1), \ldots, \phi^Y(y_T)\right)$, and $\Upsilon^Z = \left(\phi^Z(z_1), \ldots, \phi^Z(z_T)\right)$. We can now estimate the covariance operators $\widehat{\mathcal{C}}_{XY|z}$ and $\widehat{\mathcal{C}}_{YY|z}$ via Equation 10; applying KBR, we get $\widehat{\mathcal{C}}_{X|y,z} = \widehat{\mathcal{C}}_{XY|z}\left(\widehat{\mathcal{C}}_{YY|z} + \lambda I\right)^{-1}\phi^Y(y)$. We express this process with Gram matrices, using a ridge parameter $\lambda$ that goes to zero at an appropriate rate with $T$ [9]:

$$\Lambda_z = \text{diag}((G_{Z,Z} + \lambda TI)^{-1}\Upsilon^{Z*}\phi^Z(z)) \qquad (12)$$

$$\widehat{\mathcal{W}}_{X|Y,z} = \Upsilon^X(\Lambda_z G_{Y,Y} + \lambda TI)^{-1}\Lambda_z\Upsilon^{Y*} \qquad (13)$$

$$\widehat{\mu}_{X|y,z} = \widehat{\mathcal{W}}_{X|Y,z}\phi^Y(y) \qquad (14)$$

where $G_{Y,Y} \stackrel{\text{def}}{=} \Upsilon^{Y*}\Upsilon^Y$ has $(i,j)$th entry $K_Y(y_i, y_j)$, and $G_{Z,Z} \stackrel{\text{def}}{=} \Upsilon^{Z*}\Upsilon^Z$ has $(i,j)$th entry $K_Z(z_i, z_j)$. The diagonal elements of $\Lambda_z$ weight the samples, encoding the conditioning information from $z$.

## 4 RKHS EMBEDDINGS OF PSRS

We are now ready to apply Hilbert space embeddings to PSRs. For now we ignore the question of learning, and simply suppose that we are given representations of the RKHS operators described below. In Section 4.1 we show how predictive states can be represented as mean embeddings. In Section 4.2 we generalize the notion of a core set of tests and define the Hilbert space embedding of PSRs. Finally, in Section 4.3 we show how to perform filtering in our embedded PSR with Kernel Bayes' Rule. We return to learning in Section 5.

### 4.1 PREDICTIVE STATE EMBEDDINGS

We begin by defining kernels on length-$N$ sequences of test observations $\tau^\mathcal{O}$, test actions $\tau^\mathcal{A}$, and histories $h$: $K_{\mathcal{T}^\mathcal{O}}(\tau^\mathcal{O}, \tau'^\mathcal{O}) \stackrel{\text{def}}{=} \langle\phi^{\mathcal{T}^\mathcal{O}}(\tau^\mathcal{O}), \phi^{\mathcal{T}^\mathcal{O}}(\tau'^\mathcal{O})\rangle_\mathcal{F}$, $K_{\mathcal{T}^\mathcal{A}}(\tau^\mathcal{A}, \tau'^\mathcal{A}) \stackrel{\text{def}}{=} \langle\phi^{\mathcal{T}^\mathcal{A}}(\tau^\mathcal{A}), \phi^{\mathcal{T}^\mathcal{A}}(\tau'^\mathcal{A})\rangle_\mathcal{G}$, and $K_\mathcal{H}(h, h') \stackrel{\text{def}}{=} \langle\phi^\mathcal{H}(h), \phi^\mathcal{H}(h')\rangle_\mathcal{L}$. Define also the mean maps

$$\mu_{\mathcal{T}^\mathcal{A}, \mathcal{T}^\mathcal{A}, \mathcal{H}} \stackrel{\text{def}}{=} \mathbb{E}\left[\phi^{\mathcal{T}^\mathcal{A}}(\tau^\mathcal{A}) \otimes \phi^{\mathcal{T}^\mathcal{A}}(\tau^\mathcal{A}) \otimes \phi^\mathcal{H}(\mathcal{H}_t)\right] \quad (15)$$

$$\mu_{\mathcal{T}^\mathcal{O}, \mathcal{T}^\mathcal{A}, \mathcal{H}} \stackrel{\text{def}}{=} \mathbb{E}\left[\phi^{\mathcal{T}^\mathcal{O}}(\tau^\mathcal{O}) \otimes \phi^{\mathcal{T}^\mathcal{A}}(\tau^\mathcal{A}) \otimes \phi^\mathcal{H}(\mathcal{H}_t)\right] \quad (16)$$

which correspond to operators $\mathcal{C}_{\mathcal{T}^\mathcal{A}, \mathcal{T}^\mathcal{A}, \mathcal{H}}$ and $\mathcal{C}_{\mathcal{T}^\mathcal{O}, \mathcal{T}^\mathcal{A}, \mathcal{H}}$. We now take our PSR state to be the conditional embedding operator which predicts test observations from test actions:

$$\mathcal{S}(h_t) = \mathcal{W}_{\mathcal{T}^\mathcal{O}|\mathcal{T}^\mathcal{A}, h_t} = \mathcal{C}_{\mathcal{T}^\mathcal{O}, \mathcal{T}^\mathcal{A}|h_t}\mathcal{C}_{\mathcal{T}^\mathcal{A}, \mathcal{T}^\mathcal{A}|h_t}^{-1} \qquad (17)$$

where $\mathcal{C}_{\mathcal{T}^\mathcal{O}, \mathcal{T}^\mathcal{A}|h_t} = \mathcal{C}_{\mathcal{T}^\mathcal{O}, \mathcal{T}^\mathcal{A}, \mathcal{H}}\mathcal{C}_{\mathcal{H}, \mathcal{H}}^{-1}\phi^\mathcal{H}(h_t)$ and $\mathcal{C}_{\mathcal{T}^\mathcal{A}, \mathcal{T}^\mathcal{A}|h_t} = \mathcal{C}_{\mathcal{T}^\mathcal{A}, \mathcal{T}^\mathcal{A}, \mathcal{H}}\mathcal{C}_{\mathcal{H}, \mathcal{H}}^{-1}\phi^\mathcal{H}(h_t)$. This definition is analogous to the finite-dimensional case, in which the PSR state is a conditional probability table instead of a conditional embedding operator.[2]

Given characteristic RKHSs, the operator $\mathcal{S}(h_t)$ uniquely encodes the predictive densities of future observation sequences given that we take future action sequences. This is an expressive representation: we can model near-arbitrary continuous-valued distributions, limited only by the existence of the conditional embedding operator $\mathcal{W}_{\mathcal{T}^\mathcal{O}|\mathcal{T}^\mathcal{A},h_t}$ (and therefore the assumptions in Section 3.3).

---

[2]In contrast to discrete PSRs, we typically consider the entire set of length-$N$ tests at once; this change makes notation simpler, and is no loss of generality since the embedding includes the information needed to predict any individual test of length up to $N$. (Computationally, we always work with sample-based representations, so the size of our set of tests doesn't matter.)

## 4.2 CORE TESTS AND HSE-PSRS

As defined above, the embedding $\mathcal{S}(h_t)$ lets us compute predictions for a special set of tests, namely length-$N$ futures. As with discrete PSRs, knowing the predictions for some tests may allow us to compute the predictions for other tests. For example, given the embedding $\mathcal{S}(h_t)$ and another set of tests $\mathcal{T}$, there may exist a function $\mathcal{F}_\mathcal{T}$ such the predictions for $\mathcal{T}$ can be computed as $\mathcal{W}_{\mathcal{T}^\mathcal{O}|\mathcal{T}^\mathcal{A},h_t} = \mathcal{F}_\mathcal{T}(\mathcal{S}(h_t))$. In this case, $\mathcal{S}(h_t)$ is a *sufficient statistic* for $\mathcal{T}$. Here, as with discrete PSRs, we focus on prediction functions that are *linear* operators; however, this assumption is mild compared to the finite case, since linear operators on infinite-dimensional RKHSs are very expressive.

A *core* set of tests is defined similarly to the discrete PSR case (Section 2): a core set is one whose embedding $\mathcal{S}(h_t)$ is a linearly sufficient statistic for the prediction of distribution embeddings of *any* finite length. Therefore, $\mathcal{S}(h_t)$ is a *state* for an embedded PSR: at each time step $t$ we remember the embedding of test predictions $\mathcal{S}(h_t)$ instead of $h_t$.

Formally, a *Hilbert space embedding of a PSR* (HSE-PSR) is a tuple $\langle K_\mathcal{O}(o,o'), K_\mathcal{A}(a,a'), N, \mathcal{F}, \mathcal{S}_o \rangle$. $K_\mathcal{O}(o,o')$ is a characteristic kernel on observations and $K_\mathcal{A}(a,a')$ is a characteristic kernel on actions. $N$ is a positive integer such that the set of length-$N$ tests is core. $\mathcal{F}$ is the set of linear operators for predicting embeddings of any-length test predictions from the length-$N$ embedding (which must exist since length-$N$ tests are a core set), and $\mathcal{S}_0 = \mathcal{S}(h_0)$ is the initial prediction for our core tests given the null history $h_0$.

## 4.3 UPDATING STATE WITH KERNEL BAYES' RULE

Given an action $a$ and an observation $o$, the HSE-PSR state update is computed using the kernel versions of conditioning and Bayes rule given in Section 3. As in Section 2, the key idea is that the set of functions $\mathcal{F}$ allows us to predict the embedding of the predictive distribution of *any* sequence of observations from the embedding of our core set of tests $\mathcal{S}(h_t)$.

The first step in updating the state is finding the embedding of tests of length $N+1$. By our assumptions, a linear operator $\mathcal{F}_{\mathcal{AOT}}$ exists which accomplishes this:

$$\mathcal{W}_{\mathcal{T}^{\mathcal{O}'},\mathcal{O}|\mathcal{T}^{\mathcal{A}'},\mathcal{A},h_t} = \mathcal{F}_{\mathcal{AOT}}\mathcal{S}(h_t) \qquad (18)$$

The second step is finding the embedding of observation likelihoods at time $t$ given actions. By our assumptions, we can do so with an operator $\mathcal{F}_{\mathcal{AO}}$:

$$\mathcal{W}_{\mathcal{O},\mathcal{O}|\mathcal{A},h_t} = \mathcal{F}_{\mathcal{AO}}\mathcal{S}(h_t) \qquad (19)$$

With the two embeddings $\mathcal{W}_{\mathcal{T}^{\mathcal{O}'},\mathcal{O}|\mathcal{T}^{\mathcal{A}'},\mathcal{A},h_t}$ and $\mathcal{W}_{\mathcal{O},\mathcal{O}|\mathcal{A},h_t}$, we can update the state given a new action and observation. First, when we choose an action $a_t$, we compute the conditional embeddings:

$$\mathcal{C}_{\mathcal{O},\mathcal{O}|h_t,a_t} = \mu_{\mathcal{O},\mathcal{O}|h_t,a_t} = \mathcal{W}_{\mathcal{O},\mathcal{O}|\mathcal{A},h_t}\phi^\mathcal{A}(a_t) \quad (20)$$

$$\mathcal{W}_{\mathcal{T}^{\mathcal{O}'},\mathcal{O}|\mathcal{T}^{\mathcal{A}'},h_t,a_t} = \mathcal{W}_{\mathcal{T}^{\mathcal{O}'}\mathcal{O}|\mathcal{T}^{\mathcal{A}'},\mathcal{A},h_t} \times_\mathcal{A} \phi^\mathcal{A}(a_t) \quad (21)$$

Here, $\times_\mathcal{A}$ specifies that we are thinking of $\mathcal{W}_{\mathcal{T}^{\mathcal{O}'}\mathcal{O}|\mathcal{T}^{\mathcal{A}'},\mathcal{A},h_t}$ as a tensor with 4 modes, one for each of $\mathcal{T}^{\mathcal{O}'}$, $\mathcal{O}$, $\mathcal{T}^{\mathcal{A}'}$, $\mathcal{A}$, and contracting along the mode $\mathcal{A}$ corresponding to the current action. Finally, when we receive the observation $o_t$, we calculate the next state by KBR:

$$\mathcal{S}(h_{t+1}) \equiv \mathcal{W}_{\mathcal{T}^{\mathcal{O}'}|\mathcal{T}^{\mathcal{A}'},h_t,a_t,o_t}$$
$$= \mathcal{W}_{\mathcal{T}^{\mathcal{O}'},\mathcal{O}|\mathcal{T}^{\mathcal{A}'},h_t,a_t} \times_\mathcal{O} \mathcal{C}^{-1}_{\mathcal{O},\mathcal{O}|h_t,a_t}\phi^\mathcal{O}(o_t) \quad (22)$$

Here, $\times_\mathcal{O}$ specifies that we are thinking of $\mathcal{W}_{\mathcal{T}^{\mathcal{O}'},\mathcal{O}|\mathcal{T}^{\mathcal{A}'},h_t,a_t}$ as a tensor with 3 modes and contracting along the mode corresponding to the current observation.

# 5 LEARNING HSE-PSRS

If the RKHS embeddings are *finite* and low-dimensional, then the learning algorithm and state update are straightforward: we estimate the conditional embedding operators directly, learn the functions $\mathcal{F}_{\mathcal{AOT}}$ and $\mathcal{F}_{\mathcal{AO}}$ by linear regression, and update our state with Bayes' rule via Eqs. 18–22. See, for example [4] or [20]. However, if the RKHS is *infinite*, e.g., if we use Gaussian RBF kernels, then it is not possible to store or manipulate HSE-PSR state directly. In Sections 5.1–5.3, we show how learn a HSE-PSR in potentially-infinite RKHSs by leveraging the "kernel trick" and Gram matrices to represent all of the required operators implicitly. Section 5.1 describes how to represent HSE-PSR states as vectors of weights on sample histories; Section 5.2 describes how to learn the operators needed for updating states; and Section 5.3 describes how to update the state weights recursively using these operators.

## 5.1 A GRAM MATRIX FORMULATION

### 5.1.1 The HSE-PSR State

We begin by describing how to represent the HSE-PSR state in Eq. 17 as a weighted combination of training data samples. Given $T$ *i.i.d.* tuples $\left\{(\tau_t^\mathcal{O}, \tau_t^\mathcal{A}, h_t)\right\}_{t=1}^T$ generated by a stochastic process controlled by a blind policy, we denote:[3]

---
[3]To get independent samples, we'd need to reset our process between samples, or run it long enough that it mixes. In practice we can use dependent samples (as we'd get from a single long trace) at the cost of reducing the convergence rate in proportion to the mixing time. We can also use dependent samples in Sec. 5.1.2 due to our careful choice of which operators to estimate.

$$\Upsilon^{\mathcal{T}^{\mathcal{O}}} = \left(\phi^{\mathcal{T}^{\mathcal{O}}}(\tau_1^{\mathcal{O}}), \ldots, \phi^{\mathcal{T}^{\mathcal{O}}}(\tau_T^{\mathcal{O}})\right) \quad (23)$$

$$\Upsilon^{\mathcal{T}^{\mathcal{A}}} = \left(\phi^{\mathcal{T}^{\mathcal{A}}}(\tau_1^{\mathcal{A}}), \ldots, \phi^{\mathcal{T}^{\mathcal{A}}}(\tau_T^{\mathcal{A}})\right) \quad (24)$$

$$\Upsilon^{\mathcal{H}} = \left(\phi^{\mathcal{H}}(h_1), \ldots, \phi^{\mathcal{H}}(h_T)\right) \quad (25)$$

and define Gram matrices:

$$G_{\mathcal{T}^{\mathcal{A}}, \mathcal{T}^{\mathcal{A}}} = \Upsilon^{\mathcal{T}^{\mathcal{A}} *} \Upsilon^{\mathcal{T}^{\mathcal{A}}} \quad (26)$$

$$G_{\mathcal{H}, \mathcal{H}} = \Upsilon^{\mathcal{H} *} \Upsilon^{\mathcal{H}} \quad (27)$$

We can then calculate an estimate of the state at time $t$ in our training sample (Eq. 17) using Eqs. 12 and 13 from the kernel Bayes' rule derivation:

$$\alpha_{h_t} = (G_{\mathcal{H}, \mathcal{H}} + \lambda TI)^{-1} \Upsilon^{\mathcal{H} *} \phi^{\mathcal{H}}(h_t) \quad (28)$$

$$\Lambda_{h_t} = \mathrm{diag}\,(\alpha_{h_t}) \quad (29)$$

$$\widehat{\mathcal{S}}(h_t) = \Upsilon^{\mathcal{T}^{\mathcal{O}}} (\Lambda_{h_t} G_{\mathcal{T}^{\mathcal{A}}, \mathcal{T}^{\mathcal{A}}} + \lambda TI)^{-1} \Lambda_{h_t} \Upsilon^{\mathcal{T}^{\mathcal{A}} *} \quad (30)$$

We will use these training set state estimates below to help learn state update operators for our HSE-PSR.

#### 5.1.2 Vectorized States

The state update operators treat states as vectors (e.g., mapping a current state to an expected future state). The state in Eq. 30 is written as an operator, so to put it in the more-convenient vector form, we want to do the infinite-dimensional equivalent of reshaping a matrix to a vector. To see how, we can look at the example of the covariance operator $\widehat{C}_{\mathcal{T}^{\mathcal{O}}, \mathcal{T}^{\mathcal{A}} | h_t}$ and its equivalent mean map vector $\hat{\mu}_{\mathcal{T}^{\mathcal{O}} \mathcal{T}^{\mathcal{A}} | h_t}$:

$$\widehat{C}_{\mathcal{T}^{\mathcal{O}}, \mathcal{T}^{\mathcal{A}} | h_t} = \Upsilon^{\mathcal{T}^{\mathcal{O}}} \Lambda_{h_t} \Upsilon^{\mathcal{T}^{\mathcal{A}} *}$$

$$\equiv \hat{\mu}_{\mathcal{T}^{\mathcal{O}} \mathcal{T}^{\mathcal{A}} | h_t} = (\Upsilon^{\mathcal{T}^{\mathcal{O}}} \star \Upsilon^{\mathcal{T}^{\mathcal{A}}}) \alpha_{h_t} \quad (31)$$

where $\star$ is the Khatri-Rao (column-wise tensor) product. The last line is analogous to Eq. 6: each column of $\Upsilon^{\mathcal{T}^{\mathcal{O}}} \star \Upsilon^{\mathcal{T}^{\mathcal{A}}}$ is a single feature vector $\phi^{\mathcal{T}^{\mathcal{O}}}(\tau_t^{\mathcal{O}}) \otimes \phi^{\mathcal{T}^{\mathcal{A}}}(\tau_t^{\mathcal{A}})$ in the joint RKHS for test observations and test actions; multiplying by $\alpha_{h_t}$ gives a weighted average of these feature vectors.

Similarly, the HSE-PSR state can be written:

$$\widehat{\mathcal{S}}(h_t) = \widehat{C}_{\mathcal{T}^{\mathcal{O}}, \mathcal{T}^{\mathcal{A}} | h_t} \widehat{C}_{\mathcal{T}^{\mathcal{A}}, \mathcal{T}^{\mathcal{A}} | h_t}^{-1}$$

$$= \Upsilon^{\mathcal{T}^{\mathcal{O}}} (\Lambda_{h_t} G_{\mathcal{T}^{\mathcal{A}}, \mathcal{T}^{\mathcal{A}}} + \lambda TI)^{-1} \Lambda_{h_t} \Upsilon^{\mathcal{T}^{\mathcal{A}} *}$$

$$\equiv (\Upsilon^{\mathcal{T}^{\mathcal{O}}} (\Lambda_{h_t} G_{\mathcal{T}^{\mathcal{A}}, \mathcal{T}^{\mathcal{A}}} + \lambda TI)^{-1} \star \Upsilon^{\mathcal{T}^{\mathcal{A}}}) \alpha_{h_t} \quad (32)$$

We can collect all the estimated HSE-PSR states, from all the histories in our training data, into one operator $\Upsilon^{\mathcal{T}^{\mathcal{O}} | \mathcal{T}^{\mathcal{A}}}$:

$$\mathcal{W}_{\mathcal{T}^{\mathcal{O}} | \mathcal{T}^{\mathcal{A}}, h_{1:T}} \equiv \Upsilon^{\mathcal{T}^{\mathcal{O}} | \mathcal{T}^{\mathcal{A}}} = \left(\widehat{\mathcal{S}}(h_1), \ldots, \widehat{\mathcal{S}}(h_T)\right) \quad (33)$$

We need several similar operators which represent lists of vectorized conditional embedding operators. Write:

$$\Upsilon^{\mathcal{T}^{\mathcal{O}'}} = \left(\phi^{\mathcal{T}^{\mathcal{O}}}(\tau_2^{\mathcal{O}}), \ldots, \phi^{\mathcal{T}^{\mathcal{O}}}(\tau_{T+1}^{\mathcal{O}})\right) \quad (34)$$

$$\Upsilon^{\mathcal{T}^{\mathcal{A}'}} = \left(\phi^{\mathcal{T}^{\mathcal{A}}}(\tau_2^{\mathcal{A}}), \ldots, \phi^{\mathcal{T}^{\mathcal{A}}}(\tau_{T+1}^{\mathcal{A}})\right) \quad (35)$$

$$\Upsilon^{\mathcal{O}} = \left(\phi^{\mathcal{O}}(o_1), \ldots, \phi^{\mathcal{O}}(o_T)\right) \quad (36)$$

$$\Upsilon^{\mathcal{A}} = \left(\phi^{\mathcal{A}}(a_1), \ldots, \phi^{\mathcal{A}}(a_T)\right) \quad (37)$$

(Our convention is that primes indicate tests shifted forward in time by one step.) Now we can compute lists of: expected next HSE-PSR states $\mathcal{W}_{\mathcal{T}^{\mathcal{O}'} | \mathcal{T}^{\mathcal{A}'}, h_{1:T}}$; embeddings of length-1 predictive distributions $\mathcal{W}_{\mathcal{O} | \mathcal{A}, h_{1:T}}$; embeddings of length-1 predictive distributions $\mathcal{W}_{\mathcal{O}, \mathcal{O} | \mathcal{A}, h_{1:T}}$; and finally extended tests $\mathcal{W}_{\mathcal{T}^{\mathcal{O}'}, \mathcal{O} | \mathcal{T}^{\mathcal{A}'}, \mathcal{A}, h_{1:T}}$. Vectorized, these become:

$$\mathcal{W}_{\mathcal{T}^{\mathcal{O}'} | \mathcal{T}^{\mathcal{A}'}, h_{1:T}} = \Upsilon^{\mathcal{T}^{\mathcal{O}'} | \mathcal{T}^{\mathcal{A}'}} \quad (38)$$

$$\mathcal{W}_{\mathcal{O} | \mathcal{A}, h_{1:T}} = \Upsilon^{\mathcal{O} | \mathcal{A}} \quad (39)$$

$$\mathcal{W}_{\mathcal{O}, \mathcal{O} | \mathcal{A}, h_{1:T}} = \Upsilon^{\mathcal{O}, \mathcal{O} | \mathcal{A}} \quad (40)$$

$$\mathcal{W}_{\mathcal{T}^{\mathcal{O}'}, \mathcal{O} | \mathcal{T}^{\mathcal{A}'}, \mathcal{A}, h_{1:T}} = \Upsilon^{\mathcal{T}^{\mathcal{O}'}, \mathcal{O} | \mathcal{T}^{\mathcal{A}'}, \mathcal{A}} \quad (41)$$

Each of these operators is computed analogously to Eqs. 32 and 33 above. The expanded columns of Eqs. 40 and 41 are of particular importance for future derivations:

$$\Upsilon_t^{\mathcal{O}, \mathcal{O} | \mathcal{A}} = \Upsilon^{\mathcal{O}, \mathcal{O}} (\Lambda_{h_t} G_{\mathcal{A}, \mathcal{A}} + \lambda TI)^{-1} \Lambda_{h_t} \Upsilon^{\mathcal{A} *} \quad (42)$$

$$\Upsilon_t^{\mathcal{T}^{\mathcal{O}'}, \mathcal{O} | \mathcal{T}^{\mathcal{A}'}, \mathcal{A}} = \Upsilon^{\mathcal{T}^{\mathcal{O}'}, \mathcal{O} | \mathcal{T}^{\mathcal{A}'}} (\Lambda_{h_t} G_{\mathcal{A}, \mathcal{A}} + \lambda TI)^{-1} \Lambda_{h_t} \Upsilon^{\mathcal{A} *} \quad (43)$$

Finally, the finite-dimensional product of any two lists of vectorized states is a Gram matrix. In particular, we need $G_{\mathcal{T}, \mathcal{T}}$ and $G_{\mathcal{T}, \mathcal{T}'}$, Gram matrices corresponding to HSE-PSR states and time-shifted HSE-PSR states:

$$G_{\mathcal{T}, \mathcal{T}} = \Upsilon^{\mathcal{T}^{\mathcal{O}} | \mathcal{T}^{\mathcal{A}} *} \Upsilon^{\mathcal{T}^{\mathcal{O}} | \mathcal{T}^{\mathcal{A}}} \quad (44)$$

$$G_{\mathcal{T}, \mathcal{T}'} = \Upsilon^{\mathcal{T}^{\mathcal{O}} | \mathcal{T}^{\mathcal{A}} *} \Upsilon^{\mathcal{T}^{\mathcal{O}'} | \mathcal{T}^{\mathcal{A}'}} \quad (45)$$

### 5.2 LEARNING THE UPDATE RULE

The above derivation shows how to get a state estimate by embedding an entire history; for a dynamical system model, though, we want to avoid remembering the entire history, and instead recursively update the state of the HSE-PSR given new actions and observations. We are now in a position to do so. We first show how to learn a feasible HSE-PSR state that we can use to initialize filtering (Section 5.2.1), and then show how to learn the prediction operators (Section 5.2.2). Finally, we show how to perform filtering with KBR (Section 5.3).

#### 5.2.1 Estimating a Feasible State

If our data consists of a single long trajectory, we *cannot* estimate the initial state $\mathcal{S}_0$, since we only see the null history once. So, instead, we will estimate an arbitrary feasible state $\mathcal{S}_*$, which is enough information to enable prediction after an initial tracking phase if we assume that our process mixes. If we have multiple trajectories, a straightforward modification of (46) will allow us to estimate $\mathcal{S}_0$ as well.

In particular, we take $\mathcal{S}_*$ to be the RKHS representation of the stationary distribution of core test predictions given the blind policy that we used to collect the data. We estimate $\mathcal{S}_*$ as the empirical average of state estimates: $\widehat{\mathcal{W}}_{\mathcal{T}^\mathcal{O}|\mathcal{T}^\mathcal{A},h_*} = \Upsilon^{\mathcal{T}^\mathcal{O}|\mathcal{T}^\mathcal{A}} \alpha_{h_*}$ where

$$\alpha_{h_*} = \frac{1}{T}\mathbf{1}_T \qquad (46)$$

### 5.2.2 Estimating the Prediction Operators

The linear prediction operators $\mathcal{F}_{\mathcal{AO}}$ and $\mathcal{F}_{\mathcal{AOT}}$ from Eqs. 18 and 19 are the critical parameters of the HSE-PSR used to update state. In particular, we note that $\mathcal{F}_{\mathcal{AO}}$ is a linear mapping from $\mathcal{W}_{\mathcal{T}^\mathcal{O}|\mathcal{T}^\mathcal{A},h_t}$ to $\mathcal{W}_{\mathcal{O}|\mathcal{A},h_t}$ and $\mathcal{F}_{\mathcal{AOT}}$ is a linear mapping from $\mathcal{W}_{\mathcal{T}^\mathcal{O}|\mathcal{T}^\mathcal{A},h_t}$ to $\mathcal{W}_{\mathcal{T}^\mathcal{O},\mathcal{O}|\mathcal{T}^\mathcal{A},\mathcal{A},h_t}$. So, we estimate these prediction operators by kernel ridge regression:

$$\widehat{\mathcal{F}}_{\mathcal{AO}} = \Upsilon^{\mathcal{O},\mathcal{O}|\mathcal{A}} \left(\Upsilon^{\mathcal{T}^\mathcal{O}|\mathcal{T}^\mathcal{A}}\right)^\dagger_{\lambda T} \qquad (47)$$

$$\widehat{\mathcal{F}}_{\mathcal{AOT}} = \Upsilon^{\mathcal{T}^\mathcal{O},\mathcal{O}|\mathcal{T}^\mathcal{A},\mathcal{A}} \left(\Upsilon^{\mathcal{T}^\mathcal{O}|\mathcal{T}^\mathcal{A}}\right)^\dagger_{\lambda T} \qquad (48)$$

These operators are (possibly) infinite-dimensional, so we never actually build them; instead, we show how to use Gram matrices to apply these operators implicitly.

### 5.3 GRAM MATRIX STATE UPDATES

We now apply kernel Bayes' rule to *update* state given a new action and observation, i.e., to implement Eqs. 18–22 via Gram matrices. We start from the current weight vector $\alpha_t$, which represents our current state $\mathcal{S}(h_t) = \Upsilon^{\mathcal{T}^{\mathcal{O}'}|\mathcal{T}^{\mathcal{A}'}} \alpha_t$.

Predicting forward in time means applying Eqs. 47 and 48 to state. We do this in several steps. First we apply the regularized pseudoinverse in Eqs. 47 and 48, which can be written in terms of Gram matrices:

$$\left(\Upsilon^{\mathcal{T}^\mathcal{O}|\mathcal{T}^\mathcal{A}}\right)^\dagger_{\lambda T} = \Upsilon^{\mathcal{T}^\mathcal{O}|\mathcal{T}^\mathcal{A}*} \left(\Upsilon^{\mathcal{T}^\mathcal{O}|\mathcal{T}^\mathcal{A}} \Upsilon^{\mathcal{T}^\mathcal{O}|\mathcal{T}^\mathcal{A}*} + \lambda TI\right)^{-1}$$

$$= (G_{\mathcal{T},\mathcal{T}} + \lambda TI)^{-1} \Upsilon^{\mathcal{T}^\mathcal{O}|\mathcal{T}^\mathcal{A}*} \qquad (49)$$

Applying Eq. 49 to the state $\Upsilon^{\mathcal{T}^{\mathcal{O}'}|\mathcal{T}^{\mathcal{A}'}} \alpha_t$ results in

$$\hat{\alpha}_t = (G_{\mathcal{T},\mathcal{T}} + \lambda TI)^{-1} \Upsilon^{\mathcal{T}^\mathcal{O}|\mathcal{T}^\mathcal{A}*} \Upsilon^{\mathcal{T}^{\mathcal{O}'}|\mathcal{T}^{\mathcal{A}'}} \alpha_t$$

$$= (G_{\mathcal{T},\mathcal{T}} + \lambda TI)^{-1} G_{\mathcal{T},\mathcal{T}'} \alpha_t \qquad (50)$$

Here the weight vector $\hat{\alpha}_t$ allows us to predict the extended tests at time $t$ conditioned on actions and observations up to time $t-1$. That is, from Eqs. 47, 48 and 50 we can write the estimates of Eqs. 18 and 19:

$$\mathcal{F}_{\mathcal{AO}} \mathcal{S}(h_t) = \Upsilon^{\mathcal{O},\mathcal{O}|\mathcal{A}} \hat{\alpha}_t$$

$$\mathcal{F}_{\mathcal{AOT}} \mathcal{S}(h_t) = \Upsilon^{\mathcal{T}^\mathcal{O},\mathcal{O}|\mathcal{T}^\mathcal{A},\mathcal{A}} \hat{\alpha}_t$$

And, from Eqs. 42 and 43 we see that

$$\Upsilon^{\mathcal{O},\mathcal{O}|\mathcal{A}} \hat{\alpha}_t = \sum_{i=1}^T [\hat{\alpha}_t]_i \, \Upsilon^{\mathcal{O},\mathcal{O}} (\Lambda_{h_i} G_{\mathcal{A},\mathcal{A}} + \lambda TI)^{-1} \Lambda_{h_i} \Upsilon^{\mathcal{A}*} \quad (51)$$

$$\Upsilon^{\mathcal{T}^{\mathcal{O}'},\mathcal{O}|\mathcal{T}^{\mathcal{A}'},\mathcal{A}} \hat{\alpha}_t$$

$$= \sum_{i=1}^T [\hat{\alpha}_t]_i \, \Upsilon^{\mathcal{T}^{\mathcal{O}'},\mathcal{O}|\mathcal{T}^{\mathcal{A}'}} (\Lambda_{h_i} G_{\mathcal{A},\mathcal{A}} + \lambda TI)^{-1} \Lambda_{h_i} \Upsilon^{\mathcal{A}*} \quad (52)$$

After choosing action $a_t$, we can condition the embedded tests by right-multiplying Eqs. 51 and 52 by $\phi^\mathcal{A}(a_t)$. We do this by first collecting the common part of Eqs. 51 and 52 into a new weight vector $\alpha_t^a$:

$$\alpha_t^a = \sum_{i=1}^T [\hat{\alpha}_t]_i \, (\Lambda_{h_i} G_{\mathcal{A},\mathcal{A}} + \lambda TI)^{-1} \Lambda_{h_i} \Upsilon^{\mathcal{A}*} \phi^\mathcal{A}(a_t) \quad (53)$$

The estimated conditional embeddings (Eqs. 20–21) are therefore:

$$\widehat{\mathcal{C}}_{\mathcal{O},\mathcal{O}|h_t,a_t} = \Upsilon^{\mathcal{O},\mathcal{O}} \alpha_t^a$$

$$\widehat{\mathcal{W}}_{\mathcal{T}^{\mathcal{O}'},\mathcal{O}|\mathcal{T}^{\mathcal{A}'},h_t,a_t} = \Upsilon^{\mathcal{T}^{\mathcal{O}'}\mathcal{O}|\mathcal{T}^{\mathcal{A}'}} \alpha_t^a$$

Or, given a diagonal matrix with the weights $\alpha_t^a$ along the diagonal, $\Lambda_t^a = \mathrm{diag}(\alpha_t^a)$, the estimated conditional embeddings can be written:

$$\widehat{\mathcal{C}}_{\mathcal{O},\mathcal{O}|h_t,a_t} = \Upsilon^{\mathcal{O}} \Lambda_t^a \Upsilon^{\mathcal{O}*} \qquad (54)$$

$$\widehat{\mathcal{W}}_{\mathcal{T}^{\mathcal{O}'},\mathcal{O}|\mathcal{T}^{\mathcal{A}'},h_t,a_t} = \Upsilon^{\mathcal{T}^{\mathcal{O}'}|\mathcal{T}^{\mathcal{A}'}} \Lambda_t^a \Upsilon^{\mathcal{O}*} \qquad (55)$$

Given a new observation $o_t$, we apply KBR (Eq. 22):

$$\alpha_t^{ao} = (\Lambda_t^a G_{\mathcal{O},\mathcal{O}} + \lambda TI)^{-1} \Lambda_t^a \Upsilon^{\mathcal{O}*} \phi^\mathcal{O}(o_t) \qquad (56)$$

Finally, given the coefficients $\alpha_t^{ao}$, the HSE-PSR state at time $t+1$ is:

$$\widehat{\mathcal{S}}(h_t) = \widehat{\mathcal{W}}_{\mathcal{T}^{\mathcal{O}'}|\mathcal{T}^{\mathcal{A}'},h_{t+1}} = \Upsilon^{\mathcal{T}^{\mathcal{O}'}|\mathcal{T}^{\mathcal{A}'}} \alpha_t^{ao} \qquad (57)$$

This completes the state update. The nonparametric state at time $t+1$ is represented by the weight vector $\alpha_{t+1} = \alpha_t^{ao}$. We can continue to recursively filter on actions and observations by repeating Eqs. 50–57.

## 6 PREDICTIONS

In the previous sections we have shown how to maintain the HSE-PSR state by implicitly tracking the operator $\mathcal{W}_{\mathcal{T}^\mathcal{O}|\mathcal{T}^\mathcal{A}}$. However, the goal of modeling a stochastic process is usually to make *predictions*, i.e., reason about properties of future observations. We can do so via mean embeddings: for example, given the state after some history $h$, $\mathcal{W}_{\mathcal{T}^\mathcal{O}|\mathcal{T}^\mathcal{A},h}$, we can fill in a sequence of test actions to find the mean embedding of the distribution over test observations:

$$\mu_{\mathcal{T}^\mathcal{O}|h,a_{1:M}} = \mathcal{W}_{\mathcal{T}^\mathcal{O}|\mathcal{T}^\mathcal{A},h} \phi^{\mathcal{T}^\mathcal{A}}(a_{1:M}) \qquad (58)$$

As is typical with mean embeddings, we can now predict the expected value of any function $f$ in our RKHS:

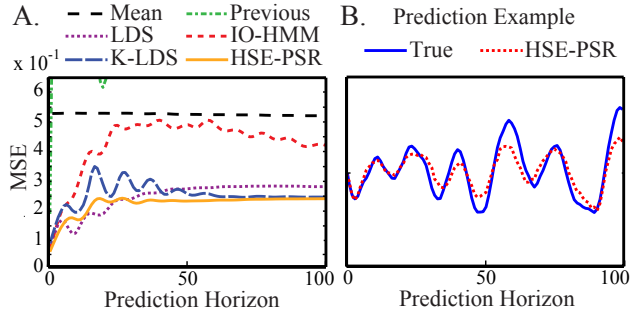

Figure 1: Synthetic data prediction performance. (A) Mean Squared Error for prediction with different estimated models. Each model was evaluated 1000 times; see text for details. (B) Example of the HSE-HMM's predicted observations given a sequence of 100 control inputs. As expected, the prediction is very accurate at the current time-step but degrades over time.

$$\mathbb{E}[f(o_{1:M}) \mid h, a_{1:M}] = \langle f, \mu_{\mathcal{T}^{\mathcal{O}} \mid h, a_{1:M}} \rangle \qquad (59)$$

The range of predictions we can make therefore depends on our RKHS. For example, write $\pi_{ij}(o_{1:M})$ for the function which extracts the $i$th coordinate of the $j$th future observation. If these coordinate projections are in our RKHS, we can compute $\mathbb{E}[(o_j)_i \mid h, a_{1:M}]$ as the inner product of $\mu_{\mathcal{T}^{\mathcal{O}} \mid h, a_{1:M}}$ with $\pi_{ij}$. (Coordinate projection functions are present, for example, in the RKHS for a polynomial kernel, or in the RKHS for a Gaussian kernel on any compact subset of a real vector space.) Or, if our RKHS contains an indicator function for a region $A$, we can predict the probability that the future observations fall in $A$.

Sometimes the desired function is absent from our RKHS. In this case, we can learn an approximation from our training data by kernel linear regression. This approximation has a particularly simple and pleasing form: we compute $f_s = f(o_{s:s+M-1})$ at each training time point $s$, collect these $f_s$ into a single vector $f$, and predict $\mathbb{E}[f(o_{1:M}) \mid h, a_{1:M}] = f^\top \alpha_h$, where $\alpha_h$ is the vector of weights representing our state after history $h$. In the experiments in Section 7 below, we use this trick to evaluate the expected next observation.

## 7 EXPERIMENTS

### 7.1 SYNTHETIC DATA

First we tested our algorithm on a benchmark synthetic nonlinear dynamical system [21, 22]:

$$\begin{aligned}
\dot{x}_1(t) =\ & x_2(t) - 0.1 \cos(x_1(t))\left(5x_1(t) - 4x_1^3(t) + x_1^5(t)\right) \\
& - 0.5 \cos(x_1(t))\, u(t), \\
\dot{x}_2(t) =\ & -65 x_1(t) + 50 x_1^3(t) - 15 x_1^5(t) - x_2(t) - 100 u(t), \\
y(t) =\ & x_1(t)
\end{aligned}$$

The output is $y$; the policy for the control input $u$ is zero-order hold white noise, uniformly distributed between $-0.5$ and $0.5$. We collected a single trajectory of 1600 observations and actions at 20Hz, and split it into 500 training and 1200 test data points.

For each model, discussed below, we filtered for 1000 different extents $t_1 = 101, \ldots, 1100$, then predicted the system output a further $t_2$ steps in the future, for $t_2 = 1, \ldots, 100$. We averaged the squared prediction error over all $t_1$; results are plotted in Figure 1(A).

We trained a HSE-PSR using the algorithm described in Section 5 with Gaussian RBF kernels and tests and histories consisting of 10 consecutive actions and observations. The bandwidth parameter of the Gaussian RBF kernels is set with the "median trick." For comparison, we learned several additional models with parameters set to maximize each model's performance: a 5-dimensional nonlinear model using a kernelized version of linear system identification (K-LDS) [22], a 5-dimensional linear dynamical system (LDS) using a stabilized version of spectral subspace identification [23, 24] with Hankel matrices of 10 time steps; and a 50-state input-output HMM (IO-HMM) trained via EM [2], with observations and actions discretized into 100 bins. We also compared to simple baselines: the mean observation and the previous observation. The results (Figure 1(A)) demonstrate that the HSE-PSR algorithm meets or exceeds the performance of the competing models.

### 7.2 SLOT CAR

The second experiment was to model inertial measurements from a slot car (1:32 scale) racing around a track. Figure 2(A) shows the car and attached 6-axis IMU (an Intel Inertiadot), as well as the 14m track. (Song et al. [20, 10] used a similar dataset.) We collected the estimated 3D acceleration and angles of the car (observations) from the IMU as well as the velocity of the car (the control input) at 10Hz for 2500 steps. We split our data into 500 training and 2000 test data points. The control policy was designed to maximize speed—it is not blind, but our learning algorithm works well despite this fact.

For each model, we performed filtering for 1000 different extents $t_1 = 501, \ldots, 1500$, then predicted an IMU reading a further $t_2$ steps in the future, for $t_2 = 1, \ldots, 500$, using the given control signal. We averaged the squared prediction error over all $t_1$; results are plotted in Figure 2(B).

The models are: an HSE-PSR with Gaussian RBF kernels on tests and histories consisting of 150 consecutive actions and observations; a 40-dimensional nonlinear model trained by K-LDS with the same settings as our HSE-PSR; a stablized 40-dimensional LDS with Hankel matrices of 150 time steps; and a 50-state IO-HMM, with observations and actions discretized into

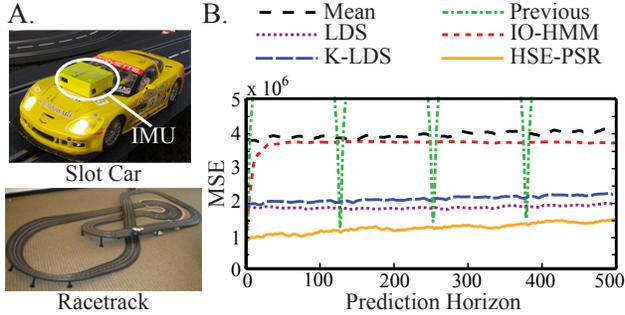

Figure 2: Slot car experiment. (A) The slot car platform: the car and IMU (top) and the racetrack (bottom). (B) Mean Squared Error for prediction with different estimated models. Each model was evaluated 1000 times; see text for details.

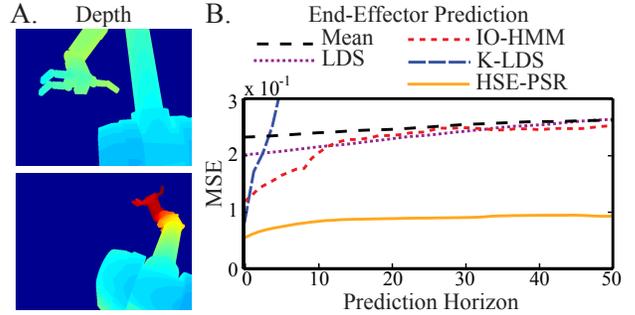

Figure 3: Robot end-effector prediction. (A) Observations consisted 640x480 pixel depth images of a robot arm. (B) Mean Squared Error (in cm) for end-effector prediction with different learned models. Each model was evaluated 400 times; see text for details.

200 bins. We again included mean and previous observation as baselines.[4] In general, the dynamical systems designed for continuous observations and controls performed well, but the HSE-PSR consistently yields the lowest RMSE.

### 7.3 ARM END-EFFECTOR PREDICTION

In the final experiment we look at the problem of predicting the 3-d position of the end-effector of a simulated Barrett WAM robot arm observed by a depth-camera. Figure 3(A) shows example depth images.

We collected 1000 successive observations of the arm motor babbling. The data set consisted of depth maps and the 3D position of the end-effector along with the joint angles of the robot arm (which we treat as the control signal). The goal was to learn a nonlinear dynamical model of the depth images and 3D locations in response to the joint angles, both to remove noise and to account for hysteresis in the reported angles. After filtering on the joint angles and depth images, we predict current and future 3D locations of the end-effector. We used the first 500 data points as training data, and held out the last 500 data points for testing the learned models.

For each model described below, we performed filtering for 400 different extents $t_1 = 51, \ldots, 450$ based on the depth camera data and the joint angles, then predicted the end effector position a further $t_2$ steps in the future, for $t_2 = 1, 2 \ldots, 50$ using just the inputs. The squared error of the predicted end-effector position was recorded, and averaged over all of the extents $t_1$ to obtain the means plotted in Figure 2(B).

We trained a HSE-PSR with Gaussian RBF kernels and tests and histories consisting of 5 consecutive actions and observations. For comparison, we learned a 100-dimensional nonlinear model using K-LDS with the same settings as our HSE-PSR, a stabilized 100-dimensional LDS with Hankel matrices of 5 time steps; and a 100-state discrete IO-HMM where observations and actions were discretized into 100 values. This is a very challenging problem and most of the approaches had difficulty making good predictions. For example, the K-LDS learning algorithm generated an unstable model and the stabilized LDS had poor predictive accuracy. The HSE-PSR yields significantly lower mean prediction error compared to the alternatives.

## 8 CONCLUSION

In this paper we attack the problem of learning a controlled stochastic process directly from sequences of actions and observations. We propose a novel and highly expressive model: *Hilbert space embeddings of predictive state representations*. This model extends discrete linear PSRs to large and continuous-valued dynamical systems. With the proper choice of kernel, HSE-PSRs can represent near-arbitrary continuous and discrete-valued stochastic processes.

HSE-PSRs also admit a powerful learning algorithm. As with ordinary PSRs, the parameters of the model can be written entirely in terms of predictive distributions of observable events. (This is in stark contrast to latent variable models, which have unobservable parameters that are usually estimated by heuristics such as EM.) Unlike previous work on continuous-valued PSRs, we do not assume that predictive distributions conform to particular parametric families. Instead, we define the HSE-PSR state as the nonparametric embedding of a conditional probability operator in a characteristic RKHS, and use recent theory developed for RKHS embeddings of distributions to derive sample-based learning and filtering algorithms.

---

[4] Like a stopped clock, the previous observation (the green dotted line) is a good predictor every 130 steps or so as the car returns to a similar configuration on the track.